\documentclass[11pt,a4paper]{article}
\usepackage[hyperref]{acl2020}
\usepackage{times}
\usepackage{latexsym}

\usepackage{graphicx}
\usepackage{fancyvrb}
\usepackage{soul}
\usepackage{alltt}

\newcommand{\unnamed}{\$}
\newcommand{\expand}{$\langle\rangle$}

\usepackage{microtype}

\aclfinalcopy 

\definecolor{qhbrackets}{RGB}{228,26,28}
\definecolor{qhconstrainkey}{RGB}{152,78,163}
\definecolor{qhconstrainvalue}{RGB}{77,175,74}
\definecolor{qhname}{RGB}{55,126,184}
\definecolor{qhop}{RGB}{255,127,0}

\newcommand{\tgn}[1]{\textcolor{qhname}{#1}} 
\newcommand{\tgv}[1]{\textcolor{qhconstrainvalue}{#1}} 
\newcommand{\tgo}[1]{\textcolor{qhop}{#1}} 
\newcommand{\tgc}[1]{\textcolor{qhconstrainkey}{#1}} 
\newcommand{\tgb}[1]{\textcolor{qhbrackets}{#1}} 

\newcommand{\sqk}[1]{\textcolor{qhname}{#1}} 
\newcommand{\sqc}[1]{\textcolor{qhconstrainkey}{#1}} 
\newcommand{\sqv}[1]{\textcolor{qhconstrainvalue}{#1}} 

\newcommand{\jsonb}[1]{\textcolor{qhbrackets}{#1}} 
\newcommand{\jsono}[1]{\textcolor{qhop}{#1}} 
\newcommand{\jsonk}[1]{\textcolor{qhname}{#1}} 

\newcommand{\yamo}[1]{\textcolor{qhop}{#1}} 
\newcommand{\yamk}[1]{\textcolor{qhname}{#1}} 
\newcommand{\yamb}[1]{\textcolor{qhbrackets}{#1}} 
 
\newcommand{\hide}[1]{}

\newcommand{\query}[1]{{\small\textsf{#1}}}

\title{Syntactic Search by Example}

\author{Micah Shlain$^{1,2}$ \hspace{1em} Hillel Taub-Tabib$^1$ \hspace{1em} \hspace {1em} Shoval Sadde$^1$ \hspace{1em} Yoav Goldberg$^{1,2}$ \\
$^1$ Allen Institute for AI, Tel Aviv, Israel \\
$^2$ Bar Ilan University, Ramat-Gan, Israel \\
\texttt{\{micahs,hillelt,shovals,yoavg\}@allenai.org yogo@cs.biu.ac.il} \\}

\date{}

\begin{document}
\maketitle
\begin{abstract}

We present a system that allows a user to search a large linguistically annotated corpus using syntactic patterns over dependency graphs. In contrast to previous attempts to this effect, we introduce a light-weight query language that 
does not require the user to know the details of the underlying syntactic representations, and
instead to query the corpus by providing an example sentence coupled with simple markup.
Search is performed at an interactive speed due to an efficient linguistic graph-indexing and retrieval engine. This allows for rapid exploration, development and refinement of syntax-based queries.
We demonstrate the system using queries over two corpora: the English wikipedia, and a collection of English pubmed abstracts. A demo of the wikipedia system is available at: \url{https://allenai.github.io/spike/} .
\end{abstract}

\section{Introduction}

The introduction of neural-network based models into NLP brought with it a substantial increase in syntactic parsing accuracy. We can now produce accurate
syntactically annotated corpora at scale. However, the produced representations themselves remain opaque to most  users, and require substantial linguistic expertise to use.
Patterns over syntactic dependency graphs\footnote{In this paper, we very loosely use the term ``syntactic''
to refer to a linguistically motivated graph-based annotation over a piece of
text, where the graph is directed and there is a path between any two nodes.
While this usually implies syntactic dependency trees or graphs (and indeed, our system currently indexes Enhanced English Universal Dependency graphs \cite{ud,eud}) the system can work also
with more semantic annotation schemes e.g, \cite{oepen-etal-2015-semeval},
 given the availability of an accurate enough parser for them.} can be very effective for interacting with linguistically-annotated corpora, either for linguistic retrieval or for information and relation extraction \cite{reverb, akbik, odin, cancer}. However, their use in mainstream NLP as represented in ACL and affiliated venues remain limited. We argue that this is due to the high barrier of entry associated with the application of such patterns. Our aim is to lower this barrier and allow also linguistically-na\"ive users to effectively experiment with and develop syntactic patterns. Our proposal rests on two components: 

\begin{figure*}[t!]
    \centering
    \includegraphics[width=0.9\textwidth]{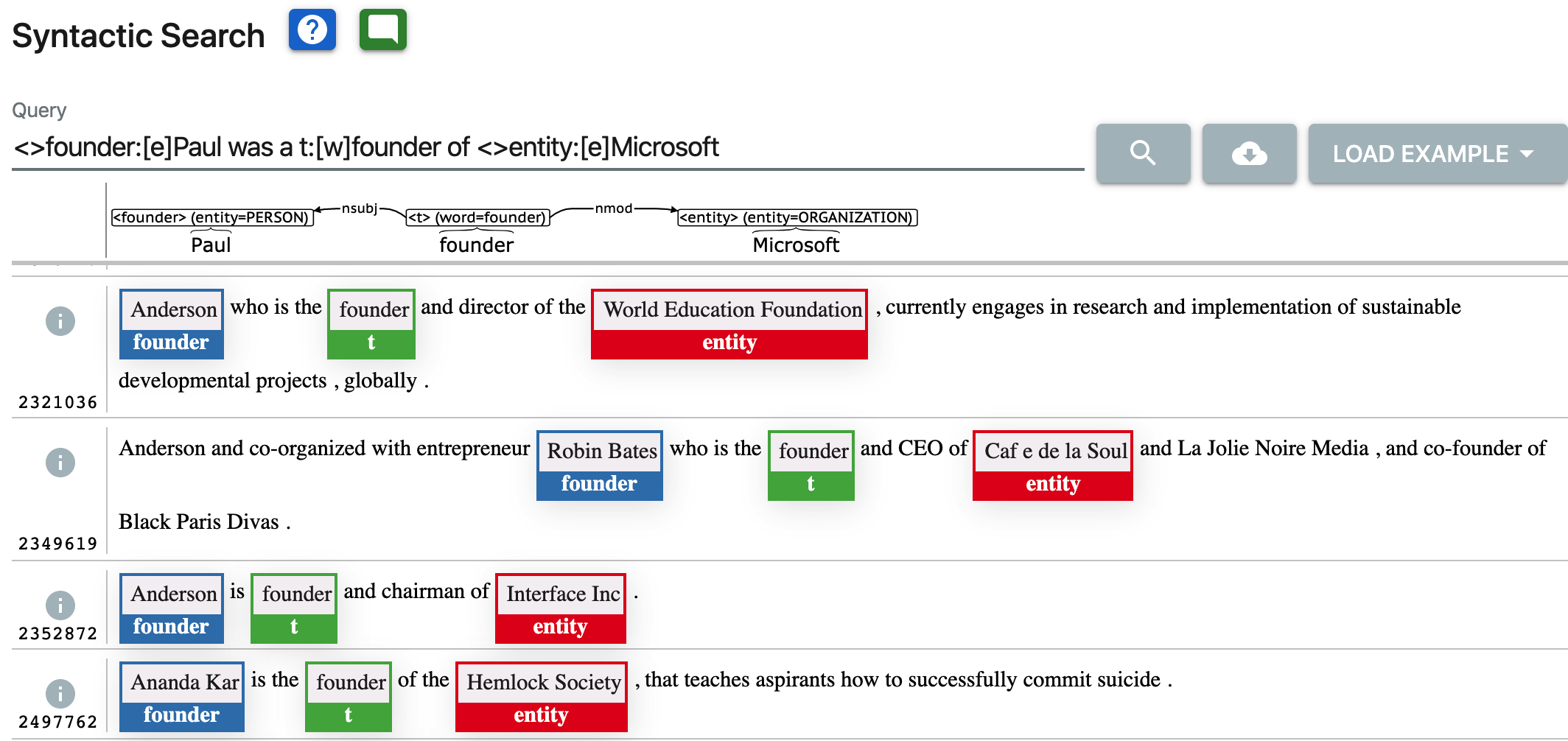}
    \caption{Syntactic Search System}
    \label{fig:main}
\end{figure*}

\noindent{\bf (1)} A light-weight query language that does not require in-depth familiarity with the underlying syntactic representation scheme, and instead lets the user specify their intent via a natural language example and lightweight markup.

\noindent{\bf (2)} A fast, near-real-time response time due to efficient indexing, allowing for rapid experimentation.

Figure \ref{fig:main} (next page) shows the interface of our web-based system. The user issued the query:\\[0.5em]
{\query{\tgb{\expand}\tgn{founder}\tgo{\textbf{:}}\tgc{[e]}Paul was a \tgn{t}\tgo{\textbf{:}}\tgc{[w]}founder of \tgb{\expand}\tgn{entity}\tgo{\textbf{:}}\tgc{[e]}Microsoft}}.\\[0.5em] 
The query specifies a sentence (\emph{Paul was a founder of Microsoft}) and three named captures: \emph{founder}, \emph{t} and \emph{entity}. The \emph{founder} and \emph{entity} captures should have the same entity-type as the corresponding sentence words (PERSON for Paul and ORGANIZATION for Microsoft, indicated by \query{[e]}), and the \emph{t} capture should have the same word form as the one in the sentence (founder) (indicated by \query{[w]}).  The syntactic relation between the captures should be the same as the one in the sentence, and the \emph{founder} and \emph{entity} captures should be \emph{expanded} (indicated by \query{\expand}).

The query is translated into a graph-based query, which is shown below the query, each graph-node associated with the query word that triggered it. The system also returned a list of matched sentences. The matched tokens for each capture group (\emph{founder}, \emph{t} and \emph{entity}) are highlighted. The user can then issue another query, browse the results, or download all the results as a tab-separated file. 

\section{Existing syntactic-query languages}
While several rich query languages over linguistic tree and graph structure exist, they require a substantial amount of expertise to use.\footnote{We focus here on systems that are based on dependency syntax, but note that many systems and query languages exist also for constituency-trees, e.g., TGREP/TGREP2, TigerSearch \cite{tigersearch}, the linguists search engine \cite{resnik-elkiss-2005-linguists}, Fangorn \cite{ghodke-bird-2012-fangorn}.}  The user needs to be familiar not only with the syntax of the query language itself, but to also be intimately familiar with the specific syntactic scheme used in the underlying linguistic annotations. For example, in Odin \cite{odin}, a dedicated language for pattern-based information extraction, the same rule as above is expressed as:
\begin{alltt}\small{
  \yamb{-} \yamk{label}\yamo{:} Person
    \yamk{type}\yamo{:} token
    \yamk{pattern}\yamo{:} \yamb{|}
     [entity="PERSON"]+
  \yamb{-} \yamk{label}\yamo{:} Organization
    \yamk{type}\yamo{:} token
    \yamk{pattern}\yamo{:} \yamb{|}
     [entity="ORGANIZATION"]+
  \yamb{-} \yamk{label}\yamo{:} founded
    \yamk{type}\yamo{:} dependency
    \yamk{pattern}\yamo{:} \yamb{|}
      trigger = [word=founded]
      founder:Person = >nsubj 
      entity:Organization = >nmod
}\end{alltt}

\noindent The Spacy NLP toolkit\footnote{\url{https://spacy.io/}} also includes pattern matcher over dependency trees,
 using JSON based syntax: 

\begin{alltt}\small{
\jsonb{[\{}\jsonk{"PATTERN"}\jsono{:} \jsonb{\{}\jsonk{"ORTH"}\jsono{:} "founder"\jsonb{\}}\jsono{,}
  \jsonk{"SPEC"}\jsono{:} \jsonb{\{}\jsonk{"NODE_NAME"}\jsono{:} "t"\jsonb{\}\}}\jsono{,}
 \jsonb{\{}\jsonk{"PATTERN"}\jsono{:} \jsonb{\{}\jsonk{"ENT_TYPE"}\jsono{:} "PERSON"\jsonb{\}\}}\jsono{,}
  \jsonk{"SPEC"}\jsono{:} \jsonb{\{}\jsonk{"NODE_NAME"}\jsono{:} "founder"\jsono{,}
           \jsonk{"NBOR_RELOP"}\jsono{:} ">nsubj"\jsono{,} 
           \jsonk{"NBOR_NAME"}\jsono{:} "t"\jsonb{\}\}}\jsono{,}
 \jsonb{\{}\jsonk{"PATTERN"}\jsono{:} \jsonb{\{}\jsonk{"ENT_TYPE"}\jsono{:} "ORGANIZATION"\jsonb{\}}\jsono{,}
  \jsonk{"SPEC"}\jsono{:} \jsonb{\{}\jsonk{"NODE_NAME"}\jsono{:} "entity"\jsono{,}
           \jsonk{"NBOR_RELOP"}\jsono{:} ">nmod"\jsono{,}
           \jsonk{"NBOR_NAME"}\jsono{:} "t"\jsonb{\}\}]}
}\end{alltt}

Stanford's Core-NLP package \cite{corenlp} includes a dependency matcher called \textsc{SemGrex},\footnote{\url{
https://nlp.stanford.edu/software/tregex.shtml}}
 which uses a more concise syntax:
\begin{alltt}\small{
\tgb{\{}\tgc{ner}\tgo{\textbf{:}}\tgv{PERSON}\tgb{\}}\tgo{=}\tgn{founder}
  <nsubj (\tgb{\{}\tgc{word}\tgo{\textbf{:}}\tgv{founder}\tgb{\}}\tgo{=}\tgn{t}
           >nmod \tgb{\{}\tgc{ner}\tgo{\textbf{:}}\tgv{ORG}\tgb{\}}\tgo{=}\tgn{entity})
}\end{alltt}


The dep\_search system\footnote{\url{http://bionlp-www.utu.fi/dep_search/}} from Turku university \cite{luotolahti-etal-2017-dep} is designed to provide a rich and expressive syntactic search over large parsebanks. They use a lightweight syntax and support working against pre-indexed data, though they do not support named captures of specific nodes.
\begin{alltt}\small{
  \tgn{PERSON} <nsubj \tgn{founder} >nmod \tgn{ORG}
}\end{alltt}

\noindent While the different systems vary in the verboseness and complexity of their own syntax (indeed, the Turku system's syntax is rather minimal), they all require the user to explicitly specify the dependency relations between the tokens, making it challenging and error-prone to write, read or edit. The challenge grows substantially as the complexity of the pattern increases beyond the very simple example we show here. 

Closest in spirit to our proposal, the \textsc{PropMiner} system of \citet{akbik-etal-2013-propminer} which lets the user enter a natural language sentence,  mark spans as \emph{subject}, \emph{predicate} and \emph{object}, and have a rule be generated automatically. 
However, the system is restricted to ternary subject-predicate-object patterns.
Furthermore, the generated pattern is written in a path-expression SQL variant (SerQL, \cite{serql}), which the user then needs to manually edit. For example, our query above will be translated to:
\begin{Verbatim}[commandchars=\\\{\},fontsize=\small]
\sqk{SELECT} subject, predicate, object
\sqk{FROM} {predicate.3} nsubj {subject},
     {predicate.3} nmod {object},
\sqk{WHERE} subject \sqc{POS} \sqv{“NNP”}
\sqk{AND} predicate.3 \sqc{POS} \sqv{“NN”}
\sqk{AND} object \sqc{POS} \sqv{“NNP”}
\sqk{AND} subject \sqc{TEXT} \sqv{“PAUL”}
\sqk{AND} predicate.3 \sqc{TEXT} \sqv{“founder”}
\sqk{AND} object \sqc{TEXT} \sqv{“Microsoft”}
\sqk{AND} subject \sqc{FULL_ENTITY}
\sqk{AND} object \sqc{FULL_ENTITY}
\end{Verbatim}

All these systems require the user to closely interact with linguistic concepts and explicitly specify graph-structures, posing a high barrier of entry for non-expert users. They also slow down expert users: formulating a complex query may require a few minutes.
Furthermore, many of these query languages are designed to match against a provided sentence,  and are not indexable.  This requires iterating over all sentences in the corpus attempting to match each one, requiring substantial time to obtain matches from large corpora.

 \citet{Augustinus2012ExampleBasedTQ} describe a system for syntactic search by example, which retrieves tree fragments and which is completely UI based. Our system takes a similar approach, but replaces the UI-only interface with an expressive textual query language, allowing for richer queries. We also return node matches rather than tree fragments. 

\section{Syntactic Search by Example}
We propose a substantially simplified language, that has the  minimal syntax and that does not require the user to know the underlying syntactic schema upfront (though it does not completely hide it from the user, allowing for exposure over time, and allowing control for expert users who understand the underlying syntactic annotation scheme). 

The query language is designed to be linguistically expressive, simple to use and amenable to efficient indexing and query. The simplicity and indexing requirements do come at a cost, though: we purposefully do not support some of the features available in existing languages. We expect these features to correlate with expertise.\footnote{Example of a query feature we do not support is quantification, i.e., ``nodes $a$ and $b$ should be connected via a path that includes {\em one or more} `conj' edges''.}
At the same time, we also seamlessly support expressing arbitrary sub-graphs, a task which is either challenging or impossible with many of the other systems. 

\noindent The language is based on the following principles:

\noindent{\bf(1)} The core of the query is a natural language sentence.

\noindent{\bf(2)} A user can specify the tokens of interest and constraints on them via lightweight markup.

\noindent{\bf(3)} While expert users can specify complex token constraints, effective constraints can be specified by pulling values from the query words.

The required syntactic knowledge from the user, both in terms of the syntax of the query language itself and in terms of the underlying linguistic formalism, remains minimal.

\vspace{-0.5em}
\section{Graph Query Formalism}
\vspace{-0.5em}
The language is structured around between-token relations and within-token constraints, where tokens can be \emph{captured}.

Formally, our query $G=(V,E)$ is a labeled directed graph, where each node $v_i \in V$ corresponds to a token, and a labeled edge $e = (v_i,v_j,\ell) \in E$ between the nodes corresponds to a between-token syntactic constraint. This query graph is then matched against parsed target sentences, looking for a correspondence between query nodes and target-sentence nodes that adhere to the token and edge constraints.

For example, the following graph specifies three tokens, where the first and second are connected via an `xcomp' relation, and the second and third via a `dobj' relation. The first token is unconstrained, while the second token must have the POS-tag of VB, and the third token must be the word home.

\noindent
\includegraphics[width=0.49\textwidth]{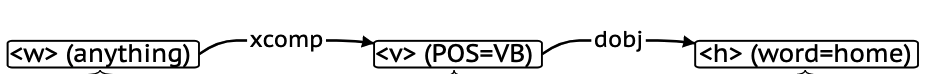}

Sentences whose syntactic graph has a subgraph that aligns to the query graph and adheres to the constraints will be considered as matches.  
Example of such matching sentences are:\\
\indent - \emph{John \underline{wanted}$_w$ to \underline{go}$_v$ \underline{home}$_h$ after lunch}. \\
\indent - \emph{It was a place she \underline{decided}$_w$ to \underline{call}$_v$ her \underline{home}$_h$}.\\
The \query{$<$w$>$}, \query{$<$v$>$} and \query{$<$h$>$} marks on the nodes denote \emph{named captures}. When matching a sentence, the sentence tokens corresponding to the graph-nodes will be bound to variables named `w', `v' and `h', in our case \texttt{\{w=wanted, v=go, h=home\}} for the first sentence and \texttt{\{w=decided, v=call, h=home\}} for the second. Graph nodes can also be unnamed, in which case they must match sentence tokens but will not bind to any variable. 
The graph structure is not meant to be specified by hand,\footnote{Indeed, we currently do not even expose a textual representation of the graph.} but rather to be inferred from the example based query language described in the next section (an example query resulting in this graph is ``{\small\query{They \tgn{w}\tgo{\textbf{:}}wanted to \tgn{v}\tgo{\textbf{:}}\tgc{[tag]}go \tgn{h}\tgo{\textbf{:}}\tgc{[word]}home}}'').

Between-token constraints correspond to labeled directed edges in the sentence's syntactic graph.

Within-token constraints correspond to properties of individual sentence tokens.\footnote{Currently supported properties are word-form (\query{word}\hide{ or \query{w}}), lemma (\query{lemma}\hide{ or \query{l}}), pos-tag (\query{tag}\hide{ or \query{t}}) or entity type (\query{entity}\hide{ or \query{e}}). Additional types can be easily added, provided that we have suitable linguistic annotators for them.}
\hide{Within-token constraints are a conjunction over disjunctions: f}For each property we specify a list of possible values (a disjunction) and if lists for several properties are provided, we require all of them to hold (a conjunction). For example, in the constraint {\small\query{tag=VBD$|$VBZ\&lemma=buy}} we look for tokens with POS-tag of either {\small\query{VBD}} or {\small\query{VBZ}}, and the lemma \emph{buy}. 
The list of possible values for a property can be specified as a pipe-separated list ({\small\query{tag=VBD$|$VBZ$|$VBN}}) or as a regular expression ({\small\query{tag=/VB[DZN]/}}). 

\hide{Formally, the structure of constraints (\texttt{Cons}) can be described by the following grammar, where \texttt{;} is used to indicate disjunction over RHS items:
\begin{verbatim}
Cons := Cons & Cons
Cons := Prop=Vals
Vals := Regexp ; Val ; Val|Vals
Prop := word ; lemma ; tag ; entity
\end{verbatim}}

\vspace{-0.5em}
\section{Example-based User-friendly Query Language}
\vspace{-0.5em}

The graph language described above is expressive enough to support many interesting queries, but it is also very tedious to specify query graphs $G$, especially for non-expert users.
We propose a simple syntax that allows to easily specify a graph query $G$ (constrained nodes connected by labeled edges) using a textual query $q$ that takes the form of an example sentence and lightweight markup.

Let $s=w_1,...,w_n$ be a proper English sentence. Let $D$ be its dependency graph, with nodes $w_i$ and labeled edges $(w_i,w_j,\ell)$. A corresponding textual query $q$ takes the form $q=q_1,...,q_n$, where each $q_i$ is either a word $q_i=w_i$, or a \emph{marked} word $q_i=m(w_i)$. A marking of a word takes the form: \query{\textbf{:}word} (unnamed capture) \query{name\textbf{:}word} (named capture) or \query{name\textbf{:}[constraints]word} , \query{\textbf{:}[constraints]word}
. 
Consider the query:\\[0.5em]
\begin{small}
\query{John \tgn{w}\tgo{\textbf{:}}wanted to \tgn{v}\tgo{\textbf{:}}\tgc{[}\tgc{tag}\tgo{=}\tgv{VB}\tgc{]} go \tgn{h}\tgo{\textbf{:}}\tgc{[}\tgc{word}\tgo{=}\tgv{home}\tgc{]} home} 
\end{small}\\[0.2em] corresponding to the above graph query.
The marked words are:\\[0.5em]
$q_2=$\query{\small{\tgn{w}\tgo{\textbf{:}}wanted}} \hfill(unconstrained, name:\query{\small{w}})\\
$q_4=$\query{\small{\tgn{v}\tgo{\textbf{:}}\tgc{[}\tgc{tag}\tgo{=}\tgv{VB}\tgc{]}go}} \hfill(cnstr:\query{\small{tag=VB}}, name:\query{\small{v}})\\
$q_5=$\query{\small{\tgn{h}\tgo{\textbf{:}}\tgc{[}\tgc{word}\tgo{=}\tgv{home}\tgc{]}home}} (cnstr:\query{\small{word=home}}, name:\query{\small{h}})\\[0.5em]
Each of these corresponds to a node $v_{q_i}$ in the query graph above.

Let $m$ be the set of marked query words, and $m^+$ be a minimal connected subgraph of $D$ that includes all the words in $m$. When translating $q$ to $G$, each marked word $w_i\in m$ is translated to a named query graph node $v_{q_i}$ with the appropriate restriction. The additional words $w_j \in m^+ \setminus m$ are translated to unrestricted, unnamed nodes $v_{q_j}$. We add a query graph edge $(v_{q_i},v_{q_j},\ell)$ for each pair in $V$ for which $(w_i,w_j,\ell) \in D$.

\noindent\textbf{Further query simplifications.}
Consider the marked word \query{\tgn{h}\tgo{\textbf{:}}\tgc{[}\tgc{word}\tgo{=}\tgv{home}\tgc{]} home}.  The constraint is redundant with the word. In such cases we allow the user to drop the value, which is then taken from the corresponding property of the query word.  
This allows us to replace the query:\\[0.5em]
\begin{small}
\query{John \tgn{w}\tgo{\textbf{:}}wanted to \tgn{v}\tgo{\textbf{:}}\tgc{[}\tgc{tag}\tgo{=}\tgv{VB}\tgc{]}go \tgn{h}\tgo{\textbf{:}}\tgc{[}\tgc{word}\tgo{=}\tgv{home}\tgc{]}home} 
\end{small}\\[0.5em]
with:\\[0.5em]
\begin{small}
\query{John \tgn{w}\tgo{\textbf{:}}wanted to \tgn{v}\tgo{\textbf{:}}\tgc{[}\tgc{tag}\tgc{]}go \tgn{h}\tgo{\textbf{:}}\tgc{[}\tgc{word}\tgc{]}home} 
\end{small}\\[0.5em]
This further drives the ``by example" agenda, as the user does not need to know what the lemma, entity-type or POS-tag of a word are in order to specify them as a constraint.
Full property names can be replaced with their shorthands \query{w,l,t,e}:\\[0.5em]
\query{John \tgn{w}\tgo{\textbf{:}}wanted to \tgn{v}\tgo{\textbf{:}}\tgc{[}\tgc{t}\tgc{]}go \tgn{h}\tgo{\textbf{:}}\tgc{[}\tgc{w}\tgc{]}home}\\[0.5em]
Finally, capture names can be omitted, in which case an automatic name is generated based on the corresponding word:\\[0.5em]
\query{John \tgo{\textbf{:}}wanted to \tgo{\textbf{:}}\tgc{[}\tgc{t}\tgc{]}go \tgo{\textbf{:}}\tgc{[}\tgc{w}\tgc{]}home} 

\noindent\textbf{Anchors.}
In some cases we want to add a node to the graph, without an explicit capture. In such cases we can use the anchor \query{\unnamed} (\query{\tgn{\unnamed}John}). These are interpreted as having a default constraint of \query{\tgc{[}\tgc{w}\tgc{]}}, which can be overriden by providing an alternative constraint (\query{\tgn{\unnamed}\tgc{[}\tgc{e}\tgc{]}John}), or an empty one (\query{\tgn{\unnamed}\tgc{[]}John}).

\noindent\textbf{Expansions}
When matching a query against a sentence the graph nodes bind to sentence words. Sometimes, we may want the match to be expanded to a larger span of the sentence. For example, when matching a word which is part of a entity, we often wish to capture the entire entity rather than the word. This is achieved by prefixing the term with the ``expansion diamond'' \query{\tgo{\expand}}.  The default behavior is to expand the match from the current word to the named entity boundary or NP-chunk that surrounds it, if it exists. We are currently investigating the option of providing additional expansion strategies.

\paragraph{Summary}
To summarize the query language from the point of view of the user: the user starts with a sentence $w_1,...,w_n$, and marks some of the words for inclusion in the query graph. For each marked word, the user may specify a name, and optional constraints. The user query is then translated to a graph query as described above. The results list highlights the words corresponding to the marked query words. The user can choose for the results to highlight entire entities rather than single words.

\vspace{-0.5em}
\section{Interactive Pattern Authoring} 
\vspace{-0.5em}
An important aspect of the system is its interactivity. Users enter queries by writing a sentence and adding markup on some words, and can then refine them following feedback from the environment, as we demonstrate with a walk-through example.

A user interested in people who obtained degrees from higher education institutions may issue the following query:\\[0.5em]
\query{\tgn{subj}\tgo{\textbf{:}}John obtained his \tgn{d}\tgo{\textbf{:}}\tgc{[}\tgc{w}\tgc{]}degree from \tgn{inst}\tgo{\textbf{:}}Harvard}\\[0.5em]
Here, the person in the ``subj'' capture and the institution in the ``inst'' capture are placeholders for items to be captured, so the user uses generic names and leaves them unconstrained. The ``degree'' (``d'') capture should match exactly, as the user specified the ``w'' constraint (exact word match).
When pressing \texttt{Enter}, the user is then shown the resulting query-graph and a result list. The user can then refine their queries based on either the query graph, the result list, or both. 
For the above query, the graph is:
\includegraphics[width=0.48\textwidth]{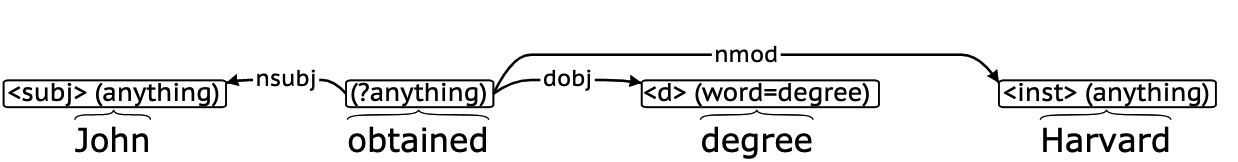}

Note that the query graph associates each graph node with the query word that triggered it. The word ``obtained'' resulted in a graph node even though it was not marked by the user as a capture. The user makes note to themselves to go back to this word later. The user also notices that the word ``from'' is not part of the query. 

Looking at the result list, things look weird:\\[0.5em]
\includegraphics[width=0.48\textwidth]{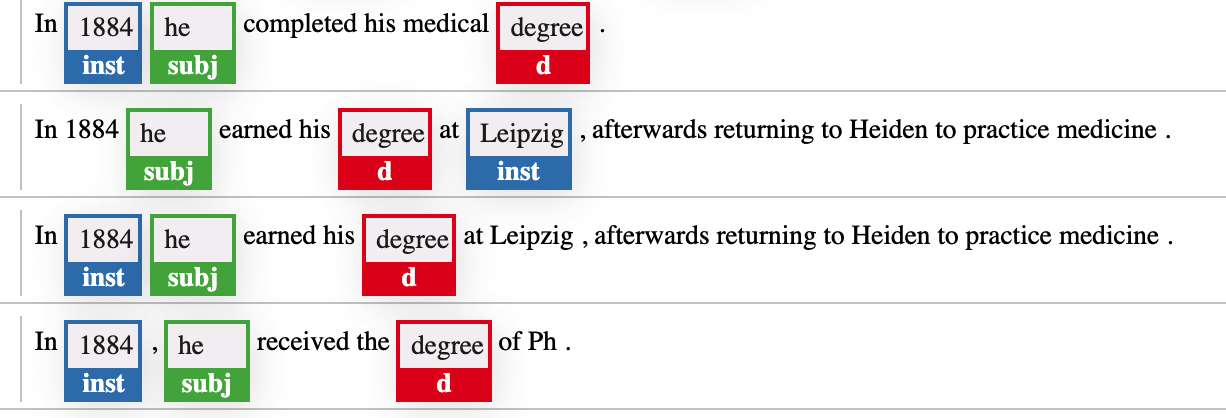}  
Maybe this is because the word \emph{from} is not in the graph?  Indeed, adding a non-capturing exact-word anchor on     ``from'' solves this issue:\\[0.5em]
\query{\tgn{subj}\tgo{\textbf{:}}John obtained his \tgn{d}\tgo{\textbf{:}}\tgc{[}\tgc{w}\tgc{]}degree \tgn{\unnamed}from \tgn{inst}\tgo{\textbf{:}}Harvard}\\[0.5em]
\includegraphics[width=0.48\textwidth]{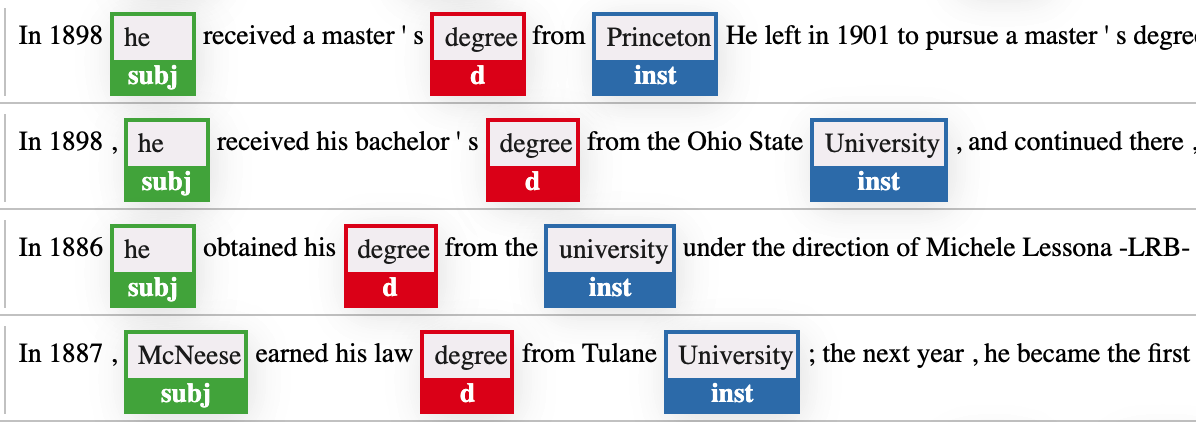}\\[0.5em]
However, the resulting list contains many non-names in the \emph{subj} capture. Trying to resolve this, the user adds an "entity-type" constraint to the \emph{subj} capture:\\[0.5em]
\query{\tgn{subj}\tgo{\textbf{:}}\tgc{[}\tgc{e}\tgc{]}John obtained his \tgn{d}\tgo{\textbf{:}}\tgc{[}\tgc{w}\tgc{]}degree \tgn{\unnamed}from \tgn{inst}\tgo{\textbf{:}}Harvard}

\noindent Note that the user didn't specify an exact type, yet the query graph correctly resolved PERSON. 

\noindent The user is interested in the full name of the person and organization, so they change from single-word capture to expanded capture, with the default expansion level (using the diamond prefix \query{\tgo{\expand}}):
\\[0.5em]
\query{\tgo{\expand}\tgn{subj}\tgo{\textbf{:}}\tgc{[}\tgc{e}\tgc{]}John obtained his \tgn{d}\tgo{\textbf{:}}\tgc{[}\tgc{w}\tgc{]}degree \tgn{\unnamed}from \tgo{\expand}\tgn{inst}\tgo{\textbf{:}}Harvard}\\[0.5em]
\includegraphics[width=0.48\textwidth]{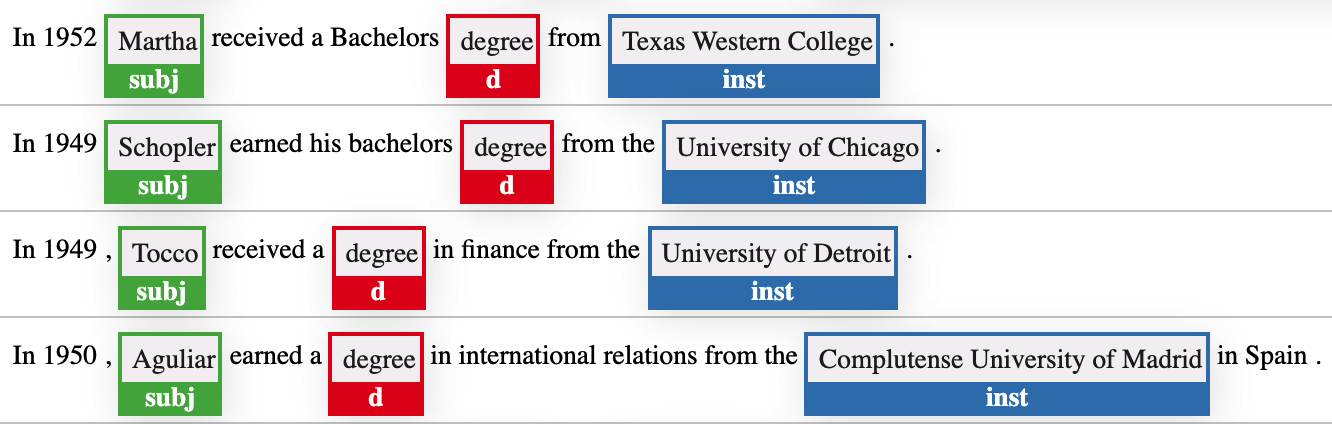}\\[0.5em]
These are the kind of results the user expected, but now they are curious about degrees obtained by females, and their representation in the Wikipedia corpus.
Adding the pronoun to the query, the user then issues the following two queries, saving the result-sets from each one as a CSV for further comparative analysis.\\[0.5em]
\query{\tgo{\expand}\tgn{subj}\tgo{\textbf{:}}\tgc{[}\tgc{e}\tgc{]}John obtained \tgn{\unnamed}his \tgn{d}\tgo{\textbf{:}}\tgc{[}\tgc{w}\tgc{]}degree \tgn{\unnamed}from \tgo{\expand}\tgn{inst}\tgo{\textbf{:}}Harvard}\\[0.5em]
\query{\tgo{\expand}\tgn{subj}\tgo{\textbf{:}}\tgc{[}\tgc{e}\tgc{]}John obtained \tgn{\unnamed}her \tgn{d}\tgo{\textbf{:}}\tgc{[}\tgc{w}\tgc{]}degree \tgn{\unnamed}from \tgo{\expand}\tgn{inst}\tgo{\textbf{:}}Harvard}\\[0.5em]
\indent Our user now worries that they may be missing some results by focusing on the word \emph{degree}. Maybe other things can be obtained from a university? The user then sets an exact-word constraint on ``Harvard'', adds a lemma constraint to ``obtain'' and clears the constraint from ``degree'':\\[0.5em]
\query{\tgo{\expand}\tgn{subj}\tgo{\textbf{:}}\tgc{[}\tgc{e}\tgc{]}John \tgo{\textbf{:}}\tgc{[}\tgc{l}\tgc{]}obtained his \tgn{d}\tgo{\textbf{:}}degree \tgn{\unnamed}from \tgo{\expand}\tgn{inst}\tgo{\textbf{:}}\tgc{[}\tgc{w}\tgc{]}Harvard}\\[0.5em]
Browsing the results, the \emph{d} capture includes words such as ``BA, PhD, MBA, certificate''. But the result list is rather short, suggesting that either \emph{Harvard} or \emph{obtain} are too restrictive. The user seeks to expand the ``obtain'' node's vocabulary, adding back the exact word constraint on ``degree'' while removing the one from ``obtain'':\\[0.5em]
\query{\tgo{\expand}\tgn{subj}\tgo{\textbf{:}}\tgc{[}\tgc{e}\tgc{]}John \tgo{\textbf{:}}\tgc{[}\tgc{]}obtained his \tgn{d}\tgo{\textbf{:}}\tgc{[}\tgc{w}\tgc{]}degree \tgn{\unnamed}from \tgo{\expand}\tgn{inst}\tgo{\textbf{:}}\tgc{[}\tgc{w}\tgc{]}Harvard}\\[0.5em]
Looking at the result list in the \emph{o} capture, the user chooses the lemmas ``receive, complete, earn, obtain, get'', adds them to the \emph{o} constraint, and removes the degree constraint.\\[0.5em]
\query{\tgo{\expand}\tgn{subj}\tgo{\textbf{:}}\tgc{[}\tgc{e}\tgc{]}John \\\tgn{o}\tgo{\textbf{:}}\tgc{[}\tgc{l}\tgo{=}\tgv{receive}\tgo{$|$}\tgv{complete}\tgo{$|$}\tgv{earn}\tgo{$|$}\tgv{obtain}\tgo{$|$}\tgv{get}\tgc{]}obtained \\ his \tgn{d}\tgo{\textbf{:}}degree \tgn{\unnamed}from \tgo{\expand}\tgn{inst}\tgo{\textbf{:}}\tgc{[}\tgc{w}\tgc{]}Harvard}\\[0.5em]
The returned result-set is now much longer, and we select additional terms for the degree slot and remove the institution word constraint, resulting in the final query:\\[0.5em]
\query{\tgo{\expand}\tgn{subj}\tgo{\textbf{:}}\tgc{[}\tgc{e}\tgc{]}John \\\tgn{o}\tgo{\textbf{:}}\tgc{[}\tgc{l}\tgo{=}\tgv{receive}\tgo{$|$}\tgv{complete}\tgo{$|$}\tgv{earn}\tgo{$|$}\tgv{obtain}\tgo{$|$}\tgv{get}\tgc{]}obtained his \tgn{d}\tgo{\textbf{:}}
\tgc{[}\tgc{w}\tgo{=}\tgv{degree}\tgo{$|$}\tgv{MA}\tgo{$|$}\tgv{BA}\tgo{$|$}\tgv{MBA}\tgo{$|$}\tgv{doctorate}\tgo{$|$}\tgv{masters}\tgo{$|$}\tgv{PhD}\tgc{]}degree \tgn{\unnamed}from \tgo{\expand}\tgn{inst}\tgo{\textbf{:}}Harvard}\\[0.5em]
The result is a list of person names earning degrees from institution, and the entire list can be downloaded as a tab-separated file which includes the named captures as well as the source sentences (over Wikipedia, this list has 6197 rows).\footnote{The list can be even more comprehensive had we selected additional degree words and obtain words, and considered also additional re-phrasings.} 
\includegraphics[width=0.48\textwidth]{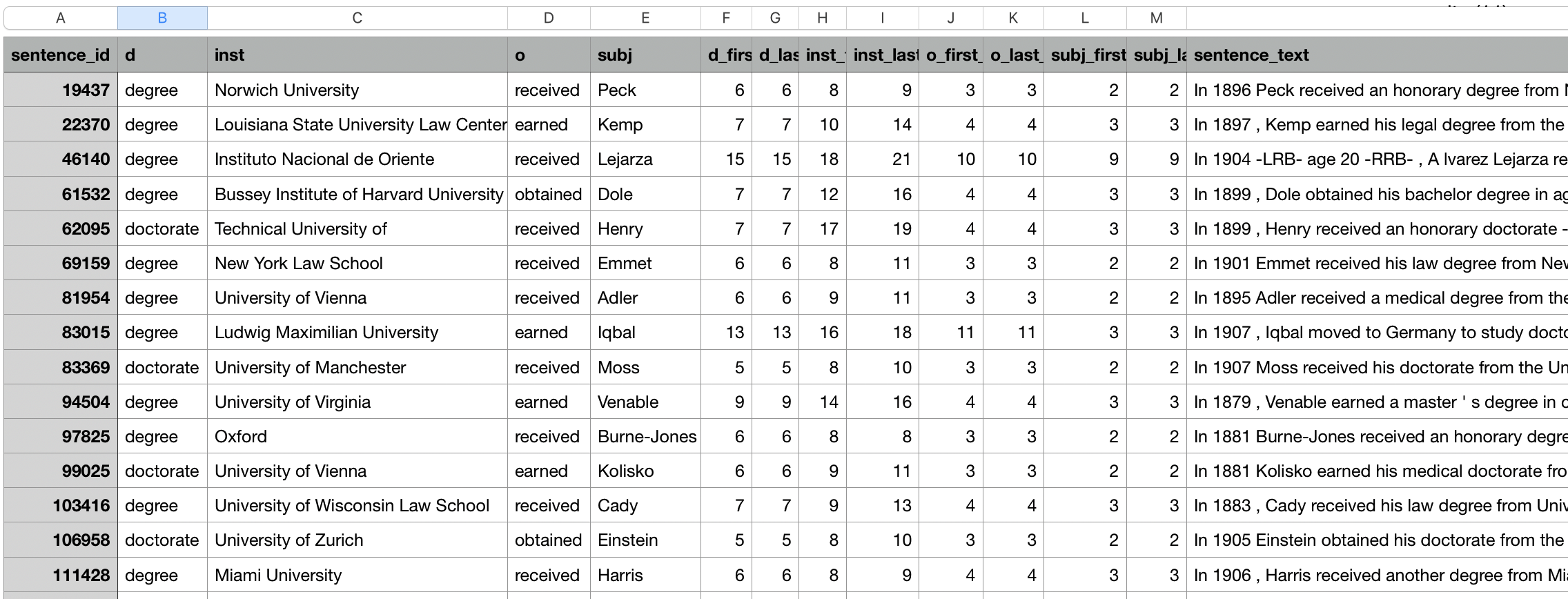}

The query can also be further refined to capture \emph{which} degree was obtained, e.g.:\\[0.5em]
\query{\tgo{\expand}\tgn{subj}\tgo{\textbf{:}}\tgc{[}\tgc{e}\tgc{]}John \tgn{o}\tgo{\textbf{:}}\tgc{[}\tgc{l}\tgo{=}\tgv{...}\tgc{]}obtained\tgb{]} his \tgn{kind}\tgo{\textbf{:}}law \tgn{d}\tgo{\textbf{:}}\tgc{[}\tgc{w}\tgo{=}\tgv{...}\tgc{]}degree \tgn{\unnamed}from \tgo{\expand}\tgn{inst}\tgo{\textbf{:}}Harvard}\\[0.5em]
capturing under \emph{kind} words like \emph{law}, \emph{chemistry}, \emph{engineering} and \emph{DLitt} but also \emph{bachelors}, \emph{masters} and \emph{graduate}.

This concludes our walk-through.

\section{Additional Query Examples}
\vspace{-0.5em}
To whet the reader's appetite, here are a sample of additional queries, showing different potential use-cases. Over wikipedia:\\
- \query{\tgn{p}\tgo{\textbf{:}}\tgc{[}\tgc{e}\tgc{]}Sam \tgn{\unnamed}\tgc{[}\tgc{l}\tgo{=}\tgv{win}\tgo{$|$}\tgv{receive}\tgc{]}won an \tgn{\unnamed}Oscar}.\\
- \query{\tgo{\expand}\tgn{p}\tgo{\textbf{:}}\tgc{[}\tgc{e}\tgc{]}Sam \tgn{\unnamed}\tgc{[}\tgc{l}\tgo{=}\tgv{win}\tgo{$|$}\tgv{receive}\tgc{]}won an \tgn{\unnamed}Oscar \tgn{\unnamed}for}\\
\indent \query{\tgo{\expand}\tgn{thing}\tgo{\textbf{:}}something}\\
- \query{\tgn{\unnamed}fish \tgn{\unnamed}such \tgn{\unnamed}as \tgo{\expand}\tgn{fish}\tgo{\textbf{:}}salmon}\\
- \query{\tgo{\expand}\tgn{hero}\tgo{\textbf{:}}\tgc{[}\tgc{t}\tgc{]}Spiderman \tgn{\unnamed}is a \tgn{\unnamed}superhero}\\
- \query{I like \tgn{kind}\tgo{\textbf{:}}coconut \tgn{\unnamed}oil}\\
- \query{\tgn{kind}\tgo{\textbf{:}}coconut \tgn{\unnamed}oil is \tgn{\unnamed}used for \tgn{purpose}\tgo{\textbf{:}}eating}\\[0.5em]
Over a pubmed corpus, annotated with the SciSpacy \cite{scispacy} pipeline:\\[0.5em]
- \query{\tgo{\expand}\tgn{x}\tgo{\textbf{:}}\tgc{[}\tgc{e}\tgc{]}aspirin \tgn{\unnamed}inhibits \tgo{\expand}\tgn{y}\tgo{\textbf{:}}thing}\\
- \query{a \tgn{\unnamed}combination of \tgo{\expand}\tgn{d1}\tgo{\textbf{:}}\tgc{[}\tgc{e}\tgc{]}aspirin and}\\ \indent\query{\tgo{\expand}\tgn{d2}\tgo{\textbf{:}}\tgc{[}\tgc{e}\tgc{]}alcohol \tgn{\unnamed}\tgo{\textbf{:}}\tgc{[}\tgc{l}\tgc{]}causes \tgo{\expand}\tgn{t}\tgo{\textbf{:}}thing}\\
- \query{\tgo{\expand}\tgn{patients}\tgo{\textbf{:}}\tgc{[}\tgc{t}\tgc{]}rats were \tgn{\unnamed}injected \tgn{\unnamed}with \tgo{\expand}\tgn{what}\tgo{\textbf{:}}drugs}

\vspace{-0.5em}
\section{Implementation Details}
\vspace{-0.5em}

The indexing is handled by Lucene.\footnote{\url{https://lucene.apache.org}} We currently use Odinson \cite{odinson},\footnote{\url{https://github.com/lum-ai/odinson/}} an
open-source Lucene-based query engine developed at Lum.ai, as a successor of Odin \cite{odin},
 that allows to index syntactic graphs and issue efficient path queries on them. We translate our queries into an Odinson path query that corresponds to a longest path in our query graph. We then iterate over the returned Odinson matches and verify the constraints that were not on the path. Conceptually, the Odinson system works by first using Lucene's reverse-index for retrieving sentences for which there is a token matching each of the specified  token-constraints, and then verifying the syntactic between-token constraints.  To improve the Lucene-query selectivity, tokens are indexed with incoming and outgoing syntactic edge label information, which is incorporated as additional token-constraints to the Lucene engine. The system easily supports millions of sentences, returning results at interactive speeds.

\vspace{-0.5em}
\section{Conclusions}
\vspace{-0.5em}
We introduce a simple query language that allows to pose complex syntax-based queries, and obtain results in an interactive speed.

A search interface over  Wikipedia sentences is available at \url{https://allenai.github.io/spike/}.
We intend to release the code as open source, as well as providing hosted open access to a PubMed-based corpus.

\section*{Acknowledgments}
We thank the team at LUM.ai and the University of Arizona, in particular Mihai Surdeanu, Marco Valenzuela-Esc\'{a}rcega, Gus Hahn-Powell and Dane Bell, for fruitful discussion and their work on the Odinson system.

This project has received funding from the Europoean Research Council (ERC) under the Europoean Union's Horizon 2020 research and innovation programme, grant agreement No. 802774 (iEXTRACT).

\bibliography{anthology,acl2020,local}
\bibliographystyle{acl_natbib}

\end{document}